\documentclass[11pt]{article}
\usepackage{coling2020/coling2020}
\usepackage{times}
\usepackage{url}
\usepackage{latexsym}
\usepackage{coling2020/comments}
\usepackage{algorithm}
\usepackage{algpseudocode}
\usepackage{amsmath}
\usepackage{empheq}
\usepackage{csquotes}

\usepackage[T2A,T1]{fontenc}
\usepackage[utf8]{inputenc}
\usepackage[russian,english]{babel}

\newauthor[S]{Simon}{green}   % \cmtS{}  \cmtS[inline]{}
\newauthor[Q]{Quan}{blue}  % \cmtQ{}   \cmtQ[inline]{}
\newauthor[M]{Mika}{orange}  % \cmtM{} \cmtM[inline]{}

\colingfinalcopy % Uncomment this line for the final submission

\title{An Unsupervised method for OCR Post-Correction and Spelling Normalisation for Finnish}

\author{Quan Duong,$^{\clubsuit}$ Mika Hämäläinen,$^{\clubsuit,\diamondsuit}$ Simon Hengchen,$^{\spadesuit}$\\
\tt{firstname.lastname@\{helsinki.fi;gu.se\}}\\
$^{\clubsuit}$University of Helsinki, $^{\diamondsuit}$Rootroo Ltd, $^{\spadesuit}$Språkbanken Text -- University of Gothenburg
}

\date{}

\begin{document}
\maketitle
\begin{abstract}
  Historical corpora are known to contain errors introduced by OCR (optical character recognition) methods used in the digitization process, often said to be degrading the performance of NLP systems. Correcting these errors manually is a time-consuming process and a great part of the automatic approaches have been relying on rules or supervised machine learning. We build on previous work on fully automatic unsupervised extraction of parallel data to train a character-based sequence-to-sequence NMT (neural machine translation) model to conduct OCR error correction designed for English, and adapt it to Finnish by proposing solutions that take the rich morphology of the language into account. Our new method shows increased performance while remaining fully unsupervised, with the added benefit of spelling normalisation. The source code and models are available on GitHub\footnote{Source Code, \url{https://github.com/ruathudo/post-ocr-correction}} and Zenodo\footnote{Trained models, \url{https://doi.org/10.5281/zenodo.4242890}}.
\end{abstract}

\section{Introduction}
\label{sec:intro}

As many people dealing with digital humanities study historical data, the problem that researchers continue to face is the quality of their digitized corpora. There have been large scale projects in the past focusing on digitizing parts of cultural history using OCR models available in those times (see \newcite{benner2003digital}, \newcite{bremer2006present}). While OCR methods have substantially developed and improved in recent times, it is still not always possible to re-OCR text. Re-OCRing documents is also not a priority when not all collections are digitised, as is the case in many national libraries. In addition, even the best OCR models will still produce errors. Noisy data is a ubiquitous problem in the digital humanities research (see e.g. \newcite{makela2020wrangling}), and tackling that problem makes it possible to answer new research questions based on old data (e.g. \newcite{saily2018explorations}).

Finnish is arguably a tremendously difficult language to tackle, due to an extremely rich morphology. This difficulty is reinforced by the limited availability of NLP tools for Finnish in general, and perhaps even more so for historical data by the fact that morphology has evolved through time -- some older inflections either do not exist anymore, or are hardly used in modern Finnish. 
As historical data comes with its own challenges, the presence of OCR errors makes the data even more inaccessible to modern NLP methods.

%What makes the problem even more difficult for Finnish is the limited availability of NLP tools that work on historical data. Also the rich morphology of the language makes the task difficult, one word can appear in many different morphological forms that may have been inflected in morphological inflections that do not exist or are hardly used in modern Finnish. As historical data comes with its own challenges, the presence of OCR errors makes the data even more inaccessible by modern NLP methods.

Obviously, this problematic situation is not unique to Finnish. There are several other languages in the world with rich morphologies and relatively poor support for both historical and modern NLP. Such is the case with most of the languages that are related to Finnish, these Uralic languages are severely endangered but have valuable historical resources in books that are not yet available in a digital format. OCR remains a problem especially for endangered languages \cite{partanen2017challenges}, although OCR quality for such languages can be improved by limiting the domain in which the OCR models are trained and used \cite{partanen2019ocr}.

Automated OCR post-correction is usually modelled as a supervised machine learning problem where a model is trained with parallel data consisting of OCR erroneous text and manually corrected text. However, we want to develop a method that can be used even in contexts where no manually annotated data is available. The most viable recent method for such a task is the one presented by \newcite{hamalainen-hengchen-2019-paft}. However, their model works only on correcting individual words, not sentences and as it focuses on English, it completely ignores the issues rising from a rich morphology. Extending their approach, we introduce a self-supervised model to automatically generate parallel data which is learned from the real OCRed text. Later, we train sequence-to-sequence NMT models on character level with the context information to correct OCR errors. The NMT models are based on the Transformer algorithm \cite{vaswani2017attention}, whose detailed comparison is demonstrated in this article.

%
% The following footnote without marker is needed for the camera-ready
% version of the paper.
% Comment out the instructions (first text) and uncomment the 8 lines
% under "final paper" for your variant of English.
% 
%\blfootnote{
    %
    % for review submission
    %
%    \hspace{-0.65cm}  % space normally used by the marker
%    Place licence statement here for the camera-ready version.
    %
    % % final paper: en-uk version 
    %
    % \hspace{-0.65cm}  % space normally used by the marker
    % This work is licensed under a Creative Commons 
    % Attribution 4.0 International Licence.
    % Licence details:
    % \url{http://creativecommons.org/licenses/by/4.0/}.
    % 
    % % final paper: en-us version 
    %
    % \hspace{-0.65cm}  % space normally used by the marker
    % This work is licensed under a Creative Commons 
    % Attribution 4.0 International License.
    % License details:
    % \url{http://creativecommons.org/licenses/by/4.0/}.
%}

\section{Related work}
\label{sec:relwork}
As more and more (digital) humanities work start to use the large-scale, digitised and OCRed collections made available by national libraries and other digitisation projects, the quality of OCR is a central point for text-based humanities research -- can one trust the output of complex NLP systems, if these are fed with bad OCR? 
Beyond the common pitfalls inherent to historical data (see \newcite{piotrowski2012natural} for a very thorough overview), some works have tried to answer the question stated above: \newcite{hill-hengchen2019OCR} use a subset of 18th-century corpus, ECCO\footnote{Eighteenth Century Collections Online, \url{https://www.gale.com/primary-sources/eighteenth-century-collections-online}} as well as its keyed-in counterpart ECCO-TCP to compare the output of common NLP tasks used in DH and conclude that OCR noise does not seem to be a large factor in quantitative analyses -- a conclusion similar to previous work by \newcite{rodriquez2012comparison} in the case of NER and to \newcite{franzini2018attributing} for authorship attribution, but in opposition to \newcite{mutuvi2018evaluating} who focus on topic modelling for historical newspapers and conclude that OCR does play a role.
More recently and still on historical newspapers, \newcite{van2020assessing} conclude that while OCR noise does have an impact, its effect widely differs between downstream tasks.

It has become obvious that OCR quality for historical texts has become central for funding bodies and collection-holding institutions alike. Reports such as the one put forward by \newcite{smith2019OCR} paint OCR initiatives, while the Library-of-Congress-commissioned report by \newcite{cordell2020OCR} underlines the importance of OCR for culturage heritage collections. 
These reports echo earlier work by, among others, \newcite{tanner2009measuring} who tackle the digitisation of British newspapers, the EU-wide IMPACT project\footnote{\url{http://www.impact-project.eu}} that gathers 26 national libraries, or \newcite{adesam2019exploring} who set out to analyse the quality of OCR made available by the Swedish language bank.

OCR post-correction has been tackled in previous work. Specifically for Finnish, \newcite{drobac2017ocr} correct the OCR of newspapers using weighted finite-state methods, while \newcite{silfverberg2015can} do the same for Finnish (and Erzya). 
Most recent approaches rely on the machine translation (MT) of ``dirty" text into ``clean" texts. These MT approaches are quickly moving from statistical MT (SMT) -- as previously used for historical text normalisation, e.g. the work by \newcite{pettersson2013smt} -- to NMT: \newcite{dong2018multi} use a word-level seq2seq NMT approach for OCR post-correction, while \newcite{hamalainen-hengchen-2019-paft}, on which we base our work, mobilised character-level NMT. Very recently, \newcite{nguyen2020neural} use BERT embeddings to improve an NMT-based OCR post-correction system on English.

\section{Experiment}
\label{sec:methods}

In this section, we describe our methods for automatically generating parallel data that can be used in a character-level NMT model to conduct OCR post-correction. In short, our method requires only a corpus with OCRed text that we want to automatically correct, a word list, a morphological analyzer and any corpus of error free text. Since we focus on Finnish only, it is important to note that such resources exist for many endangered Uralic languages as well as they have extensive XML dictionaries and FSTs available (see \cite{hamalainen2018advances}) together with a growing number of Universal Dependencies \cite{nivre2016universal} treebanks such as Komi-Zyrian \cite{lim2018multilingual}, Erzya \cite{rueter2018towards}, Komi-Permyak \cite{rueter2020questions} and North Sami \cite{sheyanova}.

\subsection{Baseline}
We design the first experiment based on previous work \cite{hamalainen-hengchen-2019-paft}, who train a character-level NMT system. To be able to train the NMT model, we need to extract the parallel data of correct words and their OCR errors. Previous research indicates that there is a strong semantic relationship between the correct word to its erroneous forms and we can generate OCR error candidates using semantic similarity. Accordingly, we trained the Word2Vec model \cite{mikolov2013efficient} on the Historical Newspaper of Finland from 1771 to 1929 using the Gensim library \cite{rehurek_lrec}. After having the Word2Vec model and its trained vocabulary, we extract the parallel data by using the Finnish morphological FST, Omorfi \cite{pirinen2015development}, provided in the UralicNLP library \cite{uralicnlp_2019} and -- following previous \newcite{hamalainen-hengchen-2019-paft} -- Levenshtein edit distance \cite{Levenshtein66Binary}. The original approach used a lemma list for English for the data extraction, but we use an FST so that we can distinguish morphological forms from OCR errors. Without the FST, different inflectional forms would also be considered to be OCR errors, which is particularly counterproductive with a highly-inflected language.

%For the dataset construction, to get the list of correct words, we use UralicNLP to lemmatize all words in the model's vocabulary, if the lemmatized words are existing in Finnish Lemma list extracted from the Finnish Wiktionary\footnote{https://fi.wiktionary.org/wiki/Wikisanakirja:Etusivu} then they are considered as the right words and be saved in the correct list. 
We build a list of correct Finnish words by lemmatisating all words in the word2vec model's vocabulary: if the lemma is present in the Finnish Wiktionary lemma list,\footnote{https://fi.wiktionary.org/wiki/Wikisanakirja:Etusivu} they are considered as correct and saved as such.
Next, for each word in this ``correct" list, we retrieve the most similar words from the Word2Vec model. Those similar words are checked to see whether they exist in the correct list or not and separated into two different groups of correct words and OCR errors. Notice that not all the words in the error list are the wrong OCR format of the given correct word, that means those words need to be filtered out. Following the paper \cite{hamalainen-hengchen-2019-paft}, we calculate the Levenshtein edit distance scores of the OCR errors to the correct word and empirically set a threshold of 4 as the maximum distance to accept as the true error form of that given word. 
As a result, for each given correct word, we have a set of correct words including the given one and a set of error words. From the two extracted groups, We do pairwise mapping to have one error word as training input and one correct word as the target output.
Finally, the parallel data is converted into a character level format before feeding it to the NMT model for training, for example: \textit{j o l e e n $\rightarrow$ j o k e e n} (``into a river"). We follow \newcite{hamalainen-hengchen-2019-paft} and use OpenNMT \cite{opennmt} with default settings, i.e. bi-directional LSTM with global attention \cite{luong2015effective}. We train for 10,000 steps and keep the last checkpoint as a baseline, which will be referred to as ``NATAS" in the remainder of this paper.

\subsection{Methods}
In the following subsections we introduce a different method to create a parallel dataset and apply a new sequence to the sequence model to train the data. 
%In the previous section, NATAS provides the parallel data by extracting the similar words from Word2Vec with a proper threshold for filtering. 
The baseline approach presented above might introduce noise when we are unable to confidently know that the error word is mapped correctly to the given correct word, especially in the case of semantically similar words that have similar lengths.
%Especially in Finnish, one word can have many inflections and it can also be a compound word. Which is a challenge for the Word2Vec model and Levenshtein edit distance to extract a quality parallel dataset. 
Another limitation of the baseline approach is the NMT model's model need for more variants to achieve correct training -- something limited by the vocabulary of the Word2Vec model, which is trained with a frequency threshold so as to provide semantically similar words. 
To solve these problems we artificially introduce OCR-like errors in a modern corpus, and thus obtain more variants of the training word pairs and less noise in the data. 
We further specialise our approach by applying the Transformer model with context and non-context words experiments instead of the default OpenNMT algorithms for training. In the next section, we detail our implementation.

\subsubsection{Dataset Construction}

For the artificial dataset, we use Yle News corpus\footnote{http://urn.fi/urn:nbn:fi:lb-2019030701} which contains more than 700 thousand articles written in Finnish from 2011 to 2018. All the articles are stored in a text file. Punctuation and characters not present in the Finnish alphabet are removed before tokenisation. After cleaning, we generate an artificial dataset by two different methods: random generator and a trained OCR error generator model.

\paragraph{Random Generator}
As previously stated, an OCR error word can be generated by a random method. In OCR text, an error normally happens when a character is misrecognized or ignored. This behavior causes some characters in the word to be missed, altered or introduced. The wrong characters will take a small ratio in the text. Thus, we introduce a strategy to produce similar errors in the modern corpus.

\begin{algorithm}[H]
\caption{Random errors generator}
\begin{algorithmic}[1]
    \Procedure{RandomError}{$Word, NoiseRate$}
    \State $Alphas$ = "abcdefghijklmnopqrstuvwxyzäåö"
    \For{\texttt{$Action$ in [delete, add, replace]}}
        \State \texttt{generate $Rand$ is a random number between 0 and 1}
        \If{$Rand < NoiseRate \times WordLength$}
            \State \texttt{Select a random character position $P$ in $Word$}
             \If{character $P$ is in $Alphas$}
                \State \texttt{Do the $Action$ on $P$ with $Alphas$}
            \EndIf
        \EndIf
    \EndFor
    \EndProcedure
    
\end{algorithmic}
\end{algorithm}

For each word in the dataset, we will manipulate errors to that word by deleting, replacing and adding characters randomly with a threshold of noise rate 0.07. The valid characters to be changed, added or removed must be in the Finnish alphabet, we do not introduce special characters as errors. The idea is that we select a random character position in the string with a probability smaller than noise rate multiplied with length of the string to restrict the percentage of errors in the word. This process is repeated for each action of deleting, replacing, adding, thus a word could either have all kinds of errors or none if the random rate is bigger than threshold. A longer word is likely to have more errors than a shorter one.

\paragraph{Trained Generator}

Similarly to the random generator, we will modify the correct word into an erroneous form, but with a different approach. Instead of pure randomness, we build a model to better simulate OCR erroneous forms. The hypothesis is that if the artificial errors introduced to words have the same pattern as the real OCRed text, it would be more effective when applying back to the real dataset. For example, the letter “i” and “l” are more likely to be misrecognized than “i” and “g” by the OCR engine. \\
To build the error generation model, we use the extracted parallel dataset from the NATAS experiment. However, the source and target for the NMT model are reversed to have correctly spelled words as the input and erroneous words as the output from the training. By trying to predict an OCR erroneous form for a given correct spelling, the model can learn an error pattern that mimics the real OCRed text. OpenNMT uses cross entropy loss by default, which causes an issue when applied to solve this problem. In our experiments, the model eventually predicted an output identical to the source because it is the most optimal way to reduce the loss. If we want to generate different output for the input, there is a need to penalize the model when having the same prediction as input. To solve the problem, we built a simple RNN translation model with GRU (gated recurrent unit) layers and a custom loss function as shown in Equation \ref{lossfunction}. The loss function is built up from cross entropy cost function in Equation \ref{crossentropy}, where $H = \{h^{(1)},..., h^{(n)}\}$ is a set of predicted outcomes from the model and $T = \{t^{(1)},..., t^{(n)}\}$ is the set of targets. We calculate normal cross entropy of predicted output $\hat{Y}$ and the labels $Y$ for finding an optimal way to mimic the target $Y$, on the other hand, the inverted cross entropy between $\hat{Y}$ and the inputs $X$ is to punish the model if the outcomes are identical to the inputs.

% \begin{figure}[H]
% \centering

% \boxed{
% cross\_entropy(A, B) = -\frac{1}{n} \sum_{i=1}^{n} a^{(i)} \ln b^{(i)} + (1 - a^{(i)}) \ln (1 - b^{(i)})
% }
% \boxed{
% loss = cross\_entropy(\hat{Y}, Y) + 1\div cross\_entropy(\hat{Y}, X)
% }
% \caption{Loss function for error generator with output $\hat{y}$, target $y$ and input $x$.}
% \label{lossfunction}
% \end{figure}

\begin{empheq}[box=\fbox]{gather}
\label{crossentropy}
cross\_entropy(H, T) = -\frac{1}{n} \sum_{i=1}^{n} t^{(i)} \ln h^{(i)} + (1 - t^{(i)}) \ln (1 - h^{(i)}) \\
\label{lossfunction}
loss = cross\_entropy(\hat{Y}, Y) + 1\div cross\_entropy(\hat{Y}, X)
\end{empheq}

The model’s encoder and decoder each have one embedding layer with 128 dimensions and one GRU layer of 512 hidden units. The input sequences are encoded to have the source’s context, this context is then passed through the decoder. For each next character of the output, the decoder concatenates the source’s context, hidden context and character’s embedded vector. The merged vectors then are passed through a linear layer to give the prediction. 
The model is trained by teacher enforcing technique with the rate 0.5. This means for the next input character, we either select the top one from the previous output or use the already known next one from the target label. 

\subsubsection{Models}

%In machine translation problem, RNN model has its cons when it comes to parallelization and long memorization \cite{bai2018empirical}. 
Parallelisation and long memorisation are weaknesses characteristic to RNNs in NMT \cite{bai2018empirical}.
Fortunately, transformers prove to be much faster (mainly due to the absence of recursion), and since they process sequences as a whole they are shown to ``remember" information better through their multi-head attention mechanism and positional embedding \cite{vaswani2017attention}. 
Transformers have been shown to be extremely efficient in various tasks (see e.g. BERT \cite{devlin2018bert}), which is why we apply this model to our problem. 
Our implementation of the transformer model is based on \cite{vaswani2017attention} and uses Pytorch framework \footnote{https://pytorch.org/}.
The model contains 3 encoder and decoder layers, each of which has 8 heads of self-attention. We also implement a learned positional encoding and use Adam \cite{kingma2014adam} as the optimizer with a static learning rate of $5\cdot10^{-4}$ which gave a better convergence compared to the default value of $0.001$ based on our experiment. Following prior work, cross entropy was again used as the loss function.

Our baseline NATAS only has fixed training samples extracted from the Word2vec model. In this experiment, we design a dynamic data loader which generates new erroneous words for every mini-batch while training, allowing the model to learn from more variants at every iteration. 
As was mentioned in the introduction, we train contextualized sequence-to-sequence character-based models. Instead of feeding a single error word to the model as the target, we combine it with the context words before and after it in sequence, as the input. We only consider the correct form of that target word as the label, and are not predicting the context words. The input from the target word in the middle and its two sides context make up a window of odd number of words. Hence, a valid window sliding over the corpus must have an odd size, for instance 3, 5, etc. The way we construct the input and gold label is presented as follows:

\begin{itemize}
\itemsep-0.4em 
\item The window size of n words is selected. The middle word is considered the target word
\item The words on left and right of the target are context words
\item The input sequence is converted in proper format, for example with window=5: \\
\texttt{<sos> l e f t <sep> c o n t e x t <ctx> f a r g e t <ctx> r i g h t <sep> c o n t e x t <eos> <pad>}, where: 
    \begin{itemize}
    \itemsep-0.3em 
    \item \texttt{<sos>} indicates the start of a sequence;
    \item \texttt{<sep>} is the separator for the context words;
    \item \texttt{<ctx>} separates left and right context with the target;
    \item \texttt{<eos>} indicates the end of a sequence;
    \item \texttt{<pad>} indicates the padding if needed for mini-batch training.
    \end{itemize}
\end{itemize}

Following the previous section, the “target” word is selected to create artificial errors in two different ways: using random generator, and a trained generator. For instance, the word “target” in the example above is modified to “farget”, and the model is trained to predict the output “target”. The gold label is also formatted in the same format, but without any context words. In this case, the label should have this form: \texttt{<sos> t a r g e t <eos>}. 
After having the pairs of input and label formatted properly, we feed them into the Transformer model with a batch size of 256 -- a balance between the speed and accuracy in our case. In this experiment, we evaluate our model with 3 different window sizes: 1, 3, and 5, with the window size of 1 as a special case: there are no context words, and the input is \texttt{<sos> f a r g e t <eos>}. 
For every window size we train with two different error generators (Random and Trained), and have thus 6 models in total. 
These models are named hereafter \textbf{TFRandW1}, \textbf{TFRandW3}, \textbf{TFRandW5}, \textbf{TFTrainW1}, \textbf{TFTrainW3}, and \textbf{TFTrainW5}, where $TF$ stands for Transformer, $Rand$ is for the random generator, $Train$ is for the trained generator and $Wn$ for a window of $n$ words.
%To make it easier to follow, we have the naming convention for 6 models as follows: \textbf{TFRandW1}, \textbf{TFRandW3}, \textbf{TFRandW5}, \textbf{TFTrainW1}, \textbf{TFTrainW3}, \textbf{TFTrainW5}, where $TF$ stands for Transformer, $Rand$ is for the random generator, $Train$ is for the trained generator and $Wn$ is for $n$ words in the window size.
We proceeded with the training until the loss converged. All models converged after around 20 epochs. The losses for the $Train$ models are $\sim0.064$ and those for $Rand$ are slightly lower, with $\sim0.059$.

\section{Evaluation}
\label{sec:evaluation}
We evaluate all proposed models and the NATAS baseline on the Ground Truth Finnish Fraktur dataset\footnote{``OCR Ground Truth Pages (Finnish Fraktur) [v1](4.8 GB)", available at \url{https://digi.kansalliskirjasto.fi/opendata}} made available by the National Library of Finland,
%The evaluation is done based on the Ground Truth Finnish Franktur dataset from Digital Collections of National Library of Finland \footnote{https://digi.kansalliskirjasto.fi/opendata}. 
a collection of 479 journal and newspaper pages from the time period 1836 - 1918 \cite{kettunen2018groundtruth}. The data format is constructed as a csv table with 471,903 lines of words or characters and there are four columns of ground truth (GT) aligned with the output coming from 3 different OCR methods TESSERACT, OLD and FR11. 

%\cmtS{I think the first line should be stronger on why we do word-error and not character-error. Any thoughts, Mika?}
%There are benchmarks for post OCR correction based on character level like in \newcite{drobac2017ocr}. However, in a word, if there is one character mis-recognized, this could lead to a different meaning.\cmtQ{Made a change} This means a high accuracy in character level does not guarantee a similar accuracy on word level which is important for human read. We analyze the result based on word level instead of character level.
Despite the existence of character-level benchmarks for OCR post-correction (e.g. \newcite{drobac2017ocr}), we elect to evaluate models on the more realistic setting of whole words. We would like to note that Finnish has very long words, and as a result this metric is actually tougher.
In the previous section, our models are trained without non-alphabet characters, so all the tokens which have non-alphabet will be removed. We also removed the blank lines which have no result from OCR. After having the ground truth and OCR text cleaned, the number of tokens for each OCR method (TESSERACT, OLD, FR11) are 458,799, 464,543 and 470,905 with accuracies of 88.29\%, 75.34\% and 79.79\% respectively. 
The OCR words will be used as input data for the evaluation of our post-correction systems.
%We will use these real OCR tokens as the input to our post-correction systems, andfix their errors to improve the accuracy. 
The translation processes apply for each OCR method separately with the input tokens formatted based on the model’s requirement. In NATAS, we used OpenNMT to translate with the default settings. In Transformer models with context, we created a sliding window over the rows of the OCRed text. For the non-context model, we only need a single token for source input. These models do the translation with beam search $k = 3$ and the highest probability sequence is chosen as the output. 
The result is shown in Table \ref{accuracypreprocessing}.

\begin{table}[H]
\centering

\begin{tabular}{|c|c|c|c|} 
\hline
Models / Targets & TESSERACT (88.29) & OLD (75.34) & FR11 (79.79)  \\ 
\hline
NATAS            & 63.35             & 61.63       & 64.95         \\ 
\hline
TFRandW1         & 69.78             & 67.33       & 71.64         \\ 
\hline
TFRandW3         & 70.02             & 67.45       & 71.69         \\ 
\hline
TFRandW5         &\textbf{71.24}     & 68.35       & 72.56         \\ 
\hline
TFTrainW1        & 70.22             & 68.30       & 72.22         \\ 
\hline
TFTrainW3        & 71.19             & 69.25       & 73.14         \\ 
\hline
TFTrainW5        & 71.24      &\textbf{69.30}       &\textbf{73.21} \\
\hline
\end{tabular}
\caption{Models accuracy on word level for all three OCR methods (\%) }
\label{accuracypreprocessing}
\end{table}

\subsection{Error Analysis}
From the result in Table \ref{accuracypreprocessing}, we can see all the models could not make any improvement on OCR text. However, there is clearly an advantage of using an artificial dataset and Transformer model for training, which has a 7\% higher accuracy compared to NATAS. After analyzing the result, we found that there are many interesting cases where the output words are considered as errors when compared to the ground truth directly but they are still correct. The difference is that the ground truth has been corrected by maintaining the historical spelling, but as our model has been trained to correct words to a modern spelling, these forms will appear as incorrect when compared directly with the ground truth. However, our models still corrected many of them right, but just happened to normalize the spelling to modern Finnish at the same time.
As examples, the word \textit{lukuwuoden} (``academic year") is normalized to \textit{lukuvuoden}, and the word \textit{kortt} (``card") is normalized to \textit{korrti}, which are the correct spellings in modern Finnish. 
So, the problem here is that many words have acquired a new spelling in modern Finnish but are seen as the wrong result if compared to the ground truth, which affects the real accuracy of our models.
In the 19th century Finnish text, the most obvious difference compared to modern Finnish is the variation of \textit{w/v}, where most of the words containing \textit{v} are written as \textit{w} in old text, whereas in modern Finnish \textit{w} is not used in any regular word. 
\newcite{kettunen2016measuring} showed in their experiments that the number of tokens containing letter \textit{w} contribute to 3.3\% of all tokens and 97.5\% of those tokens is missrecognized by FINTWOL -- a morphological analyzer. They also tried to replace the \textit{w} with \textit{v} and the unrecognized tokens decreased to 30.6\%. These numbers are significant which give us an idea to apply it on our results to get a better evaluation. 
Furthermore, there is another issue for our models when they try to make up the new words which do not exist in Finnish vocabulary. For example the word \textit{samppaajaa} is likely created from the word \textit{samppanjaa} (``of Champagne") which must be the correct one. To solve these issues, we suggested a fixing pipeline for our result:

\begin{enumerate}
  \item Check if the words exist in Finnish vocabulary using Omorfi with UralicNLP, if not then keep the OCRed words as the output.
  \item Find all words containing letter \textit{v}, replace by letter \textit{w}.
\end{enumerate}

After the processing with the strategy above, we get updated results which can be found in Tables~\ref{resultstesseract}, \ref{resultsold}, and \ref{resultsfr11}.

\begin{table}[H]
\centering

\begin{tabular}{|c|c|c|c|} 
\hline
Models    & Post processed accuracy & Error words accuracy & Correct words accuracy  \\ 
\hline
NATAS     & 74.71                   & 16.54                & 82.43                   \\ 
\hline
TFRandW1  & 80.49                   & 16.13                & 89.03                   \\ 
\hline
TFRandW3  & 80.79                   & 16.94                & 89.26                   \\ 
\hline
TFRandW5  & 81.89                   & 17.02                & 90.49                   \\ 
\hline
TFTrainW1 & 83.05                   & 17.11                & 91.79                   \\ 
\hline
TFTrainW3 & 83.96                   &\textbf{18.15}        & 92.68                   \\ 
\hline
TFTrainW5 &\textbf{84.00}           & 18.02                &\textbf{92.75}           \\
\hline
\end{tabular}
\caption{Models accuracy post-processing for Tesseract (88.29\%)}
\label{resultstesseract}
\end{table}

\begin{table}[H]
\centering

\begin{tabular}{|c|c|c|c|} 
\hline
Models    & Post processed accuracy & Error words accuracy & Correct words accuracy  \\ 
\hline
NATAS     & 71.19                   & 30.66                & 84.45                   \\ 
\hline
TFRandW1  & 75.10                   & 28.14                & 90.47                   \\ 
\hline
TFRandW3  & 75.40                   & 28.26                & 90.83                   \\ 
\hline
TFRandW5  & 76.26                   & 28.63                & 91.85                   \\ 
\hline
TFTrainW1 & 78.19                   & 35.07                & 92.30                   \\ 
\hline
TFTrainW3 &\textbf{79.26}           &\textbf{36.03}        & 93.41                   \\ 
\hline
TFTrainW5 & 79.17                   & 35.41                &\textbf{93.50}           \\
\hline
\end{tabular}
\caption{Models accuracy post-processing for OLD (75.34\%)}
\label{resultsold}
\end{table}

\begin{table}[H]
\centering

\begin{tabular}{|c|c|c|c|} 
\hline
Models    & Post processed accuracy & Error words accuracy & Correct words accuracy  \\ 
\hline
NATAS     & 75.06                   & 36.52                & 84.81                   \\ 
\hline
TFRandW1  & 79.66                   & 36.04                & 90.71                   \\ 
\hline
TFRandW3  & 80.06                   & 37.00                & 90.96                   \\ 
\hline
TFRandW5  & 81.09                   & 38.04                & 91.99                   \\ 
\hline
TFTrainW1 & 82.39                   & 43.39                & 92.26                   \\ 
\hline
TFTrainW3 &\textbf{83.50}           &\textbf{45.17}        & 93.21                   \\ 
\hline
TFTrainW5 & 83.34                   & 44.01                &\textbf{93.30}           \\
\hline
\end{tabular}
\caption{Models accuracy post-processing for FR11 (79.79\%)}
\label{resultsfr11}
\end{table}

The results in Tables \ref{resultstesseract}, \ref{resultsold} and \ref{resultsfr11} show a vast improvement for all models with the accuracy increased by 10-12\%. 
In Tesseract, where the original OCR already has a very high quality with an accuracy of 88\%, there is no gain for all models. 
The best model in this case is TFTrainW5 with 84\% accuracy. The reason for the models' worse performance is that they introduced more errors on the already correct words by OCR than fixing actual error words. 
While the ratio of fixing the error words (18.02\%) is much higher than the ratio of confounding the correct words (7.25\%), however, due to the number of correct words taking a much larger part in the corpus, the overall accuracy is decreased. %\cmtS{I don't understand this sentence} \cmtQ{You don't understand the OLD part or the previous sentence? I edited the OLD part.}
In the OLD setting with a 75\% text accuracy, 5 out of 7 models have successfully improved the accuracy of the original text. The highest number comes to TFTrainW3 which outperforms OLD by 3.92\%, and following closely is TFTrainW5 with an accuracy of 79.17\%. In OLD, we see better error words corrected (36.03\%) compared to Tesseract. The accuracy of the TFTrainW5 model for the already corrected words is also slightly higher with 93.5\% versus Tesseract 92.75\%. 
The last OCR method for evaluation is FR11 (79\%), where -- just like in OLD -- 5 out of 7 models surpass the OCR result. Again, the TFTrainW3 gives the highest number with 3.71\% improvement on the OCRed text. While the TFTrainW3 shows surprisingly good results on fixing the wrong words with 45.17\% accuracy, the TFTrainW5 performs slightly better at handling the right words. 
Common to all our proposed models, the window size of 1 somewhat unsurprisingly performs worse within both the $Rand$ and $Train$ variants.

\section{Conclusion and Future work}
\label{sec:conclusion}

In this paper, we have shown that creating and using an artificial error dataset clearly outperforms the NATAS baseline \cite{hamalainen-hengchen-2019-paft}, with a clear advantage for the $Train$ over the $Rand$ configuration. Another clear conclusion is that a larger context window results in increasing the accuracy of the models. 
Comparing the new results for all three OCR methods, we see the models are most effective with FR11 when the ratio of fixing wrong words (45.17\%) is high enough to overcome the issue of breaking the right words (6.7\%). Our methods also work very well on OLD with ability to fix 36.03\% of wrong words and handle more than 93\% of right words correctly. 
However, our models are not compelling enough to beat the accuracy achieved by Tesseract, a conclusion we see as further work.

In spite of the effectiveness of the post-correction strategy, it does not guarantee that all the words with \textit{w/v} replaced are correct, nor that UralicNLP manages to recognize all the existing Finnish words. For example: the wrong OCR word \textit{mcntoistamuotiscn} was fixed to \textit{metoistavuotisen} which is the correct one according to the gold standard, but UralicNLP has filtered it out due to not considering that is the valid Finnish word. This is true, as the first syllable \textit{kol} was dropped out due to a line break in the data, and without the line break, the word would be \textit{kolmetoistavuotisen} (``13 years old"). This means that in the future, we need to develop better strategies more suitable to OCR contexts for telling correct and incorrect words apart.

This implies that in reality the corrected cases can be higher if we don’t revert the already normalized \textit{w/v} words. In addition, if there is a better method to ensure a word is valid in Finnish, the result could be improved. Thus, our evaluation provides an overall look of how the Transformer and Trained Error Generator models with context words could improve the post OCR correction notably. Our methods also show that using artificial dataset from a modern corpus is very potential to normalize the historical text.

Importantly, we would like to underline that our method does not rely on huge amounts of hand annotated gold data, but can rather be applied for as long as one has access to an OCRed text, a vocabulary list, a morphological FST and error-free data. There are several endangered languages related to Finnish that already have these aforementioned resources in place. In the future, we are interested in trying our method out in those contexts as well.

\section*{Acknowledgments}
%QD is funded by...
SH is funded by the project \textit{Towards Computational Lexical Semantic Change Detection} supported by the Swedish Research Council (2019--2022; dnr 2018-01184). This work has been supported by the European Union Horizon 2020 research and innovation programme under grant 770299 (NewsEye).

\bibliographystyle{coling2020/coling}
\bibliography{coling2020/coling2020}

\end{document}